\documentclass{article}

\usepackage{arxiv}

\usepackage[utf8]{inputenc} 
\usepackage[T1]{fontenc}    
\usepackage{hyperref}       
\usepackage{url}            
\usepackage{booktabs}       
\usepackage{amsfonts}       
\usepackage{nicefrac}       
\usepackage{microtype}      
\usepackage{lipsum}		
\usepackage{graphicx}
\usepackage{natbib}
\usepackage{doi}
\usepackage{amsmath}
\usepackage{float}

\usepackage{subcaption}

\usepackage[utf8]{inputenc} 
\usepackage[T1]{fontenc}    
\usepackage{hyperref}       
\usepackage{nicefrac}       
\usepackage{microtype}      
\usepackage{xcolor}         
\usepackage{graphicx}

\title{\emph{gLIME}: A new graphical methodology for interpretable model-agnostic explanations}

\author{ \href{https://orcid.org/0000-0003-4451-8648}{\includegraphics[scale=0.06]{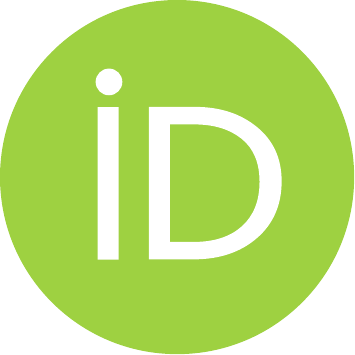}\hspace{1mm}Zoumpolia Dikopoulou}\thanks{https://www.aideas.eu/} \\
	AIDEAS OÜ,\\
	Tallinn, 10117, Estonia.\\
	\texttt{dikopoulia@gmail.com} \\
	\And
	\href{https://orcid.org/0000-0002-1090-2177}{\includegraphics[scale=0.06]{orcid.pdf}\hspace{1mm}Serafeim Moustakidis} \\
	AIDEAS OÜ,\\
	Tallinn, 10117, Estonia.\\
	\texttt{s.moustakidis@aideas.eu} \\
	
    \And
	Patrik Karlsson \\
	AIDEAS OÜ,\\
	Tallinn, 10117, Estonia.\\
	\texttt{p.karlsson@aideas.eu} \\

}

\renewcommand{\shorttitle}{\emph{gLIME}: A new graphical methodology for interpretable model-agnostic explanations }

\hypersetup{
pdftitle={gLIME: A new graphical methodology for interpretable model-agnostic explanations},
pdfsubject={q-bio.NC, q-bio.QM},
pdfauthor={David S.~Hippocampus, Elias D.~Striatum},
pdfkeywords={First keyword, Second keyword, More},
}

\begin{document}
\maketitle

\begin{abstract}

	Explainable artificial intelligence (XAI) is an emerging new domain in which a set of processes and tools allow humans to better comprehend the decisions generated by black box models. However, most of the available XAI tools are often limited to simple explanations mainly quantifying the impact of individual features to the models’ output. Therefore, human users are not able to understand how the features are related to each other to make predictions, whereas the inner workings of the trained models remain hidden. This paper contributes to the development of a novel graphical explainability tool that not only indicates the significant features of the model, but also reveals the conditional relationships between features and the inference capturing both the direct and indirect impact of features to the models’ decision.  The proposed XAI methodology, termed as gLIME, provides graphical model-agnostic explanations either at the global (for the entire dataset) or the local scale (for specific data points). It relies on a combination of local interpretable model-agnostic explanations (LIME) with graphical least absolute shrinkage and selection operator (GLASSO) producing undirected Gaussian graphical models. Regularization is adopted to shrink small partial correlation coefficients to zero providing sparser and more interpretable graphical explanations. Two well-known classification datasets (BIOPSY and OAI) were selected to confirm the superiority of gLIME over LIME in terms of both robustness and consistency/sensitivity over multiple permutations. Specifically, gLIME accomplished increased stability over the two datasets with respect to features’ importance (76\%-96\% compared to 52\%-77\% using LIME). gLIME demonstrates a unique potential to extend the functionality of the current state-of-the-art in XAI by providing informative graphically given explanations that could unlock black boxes.

\end{abstract}


\section{Introduction}
Superhuman capacity has been recently demonstrated by artificial intelligence (AI) leading to a widespread adoption of AI systems in multiple domains including healthcare, industry 4.0 and finance. However, this improved predictive performance typically comes with increased model complexity \cite{lecun2015deep}. The great majority of recent powerful machine learning algorithms are ‘black box’ approaches with the rationale behind their decision-making mechanism being hard to understand and interpret.  The ambiguity with respect to the models’ inner workings and the fact that their decisions cannot be interpreted make these systems difficult to be trusted by the end-users, especially in critical domains such as in healthcare. Meeting the need for trustworthy, fair and robust AI decisions, explainable AI (XAI) \cite{gunning2019darpa} has emerged as a new scientific field that focuses on the understanding and interpretation of AI systems’ behavior. \par
So far, XAI has been approached from different view-points with respect to the type of data (tabular, images or text) or the interpretability scale \cite{linardatos2021explainable}. As far as the interpretability scale, some techniques provide explanations for individual instances (local scale), whereas there are XAI tools or libraries used to explain the behavior of the model on the whole dataset (global scale).  Another important classification of XAI tools is related to the type of algorithms/networks in which the explainability analysis is applied. There are XAI techniques that are model specific (restricted to a specific machine learning model or to specific family of models) and other that are model-agnostic capable to be applied to any model.   \par
Post-hoc explainability refers to a specific category of XAI that encompasses techniques that explain the decisions of already trained black-box models. A considerable amount of experiments and scientific work has been devoted on the explainability of deep learning models and thus a variety of model-specific XAI tools has been proposed including DeepLIFT \cite{shrikumar2017learning}, Class Activation Maps (CAMs), first introduced in \cite{zhou2016learning}, and Grad-CAM \cite{selvaraju2017grad}. Among the post-hoc model-agnostic techniques, the local interpretable model-agnostic explanations (LIME) method \cite{ribeiro2016should} is one of the most popular methods for black-box models that generates interpretations at the local scale (for single instances). LIME is a simple but powerful technique that derives explanations utilizing simulated randomly-sampled data around the neighbourhood of an input instance. However, LIME has been proved sensitive to these randomly generated permutations leading to unstable interpretations \cite{garreau2020explaining}. Shapley Additive explanations (SHAP) \cite{lundberg2017unified} is another well-known game-theory inspired technique that estimates the importance of each feature on individual predictions, demonstrating both accuracy and consistency.  Overall, all the aforementioned model-agnostic techniques, including SHAP, do not take feature dependence into account and in some cases produce non-intuitive feature importance values. \par
Current post-hoc model-agnostic XAI techniques are limited to a very specific view-point of XAI where feature importance values are calculated and visualized with bar graphs or other similar visualization tools. To the best of our knowledge, none of the available techniques is capable of identifying the relationships between features and the possible indirect effect of features to the models’ output. This paper contributes to the development of a novel graphical explainability tool that not only indicates the significant features of the model, but also reveals the conditional relationships among features capturing and visualizing both the direct and indirect impact of features to the models’ decision.  The proposed XAI methodology, termed as gLIME, visualizes model-agnostic explanations with intuitive interpretable graphs at either the global (explanations for the entire dataset) or the local scale (explanations for specific data points). It relies on a combination of local interpretable model-agnostic explanations (LIME) with graphical least absolute shrinkage and selection operator (GLASSO) \cite{epskamp2018tutorial,meinshausen2006high} producing undirected Gaussian graphical models. In this paper we demonstrate the effectiveness of gLIME at the local scale and we compare it with LIME that shares similar characteristics. An extensive experimental analysis has been performed using two well-known classification datasets to confirm the superiority of gLIME over LIME in terms of both robustness and consistency/sensitivity over multiple permutations. \par
This paper is organized as follows. Section 2 gives an overview of the proposed gLIME methodology presenting its main characteristics and features. Results on multiple experiments are provided in Section 3, whereas conclusions are drawn in Section 4.

\section{Methods}
\label{gen_inst}
The fundamental idea behind the gLIME algorithm is focused on the local model-agnostic explainability and interpretability increasing the trust and the stability of the produced model. This signifies that the proposed methodology attempts to explain and interpret individual predictions of a model-agnostic method, in which gLIME can be applied to any supervised regression or classification model. This algorithm consists of two parts; the first one prepares the data by generating additional data around the selected observation and the second part, estimates a graph model. Generally, a graph or a network G is an abstract model which represents complex phenomena, and it consists of two components, nodes and edges \cite{dorogovtsev2013evolution}. Nodes or vertices (V) represent entities (features) and they are visualized as circles; while, edges or links (E) connect the nodes between them representing their relationships. When graph models are estimated from data structures, the edges may represent unknown statistical relationships such as: correlations, covariances, partial correlations, regression coefficients, factor loadings, etc. \cite{hevey2018network}. \par
Therefore, gLIME incorporates graphs to explain and interpret the decisions of trained models for three main reasons: i) to identify the relevant features that affect the model's prediction, ii) highlight hidden and/or important relationships among features quantifying the strength of the relationships and iii) determine significant path routes presenting how the information flows from one node to the end node (the predicted feature) traversing the intermediate significant nodes that can affect indirectly the predicted output. \par
Below the steps of gLIME are described. Before dividing the dataset into training and testing, the dataset must be clean from missing values. Then, a selected ML model is applied to the training data for a classification or regression problem. In our experiment, Non linear Support Vector Machines (SVMs) were utilized to perform the classification tasks. From the testing data, an observation is chosen and it is permuted m times to create replicated feature data with minor value variations. Next, a similarity distance measure is applied to calculate the distance between the initial observation and the permuted observations. Particularly, the distance measure is converted to a similarity value with the use of an exponential kernel which by default is adjusted to 0.75 times the square root of the number of features. Afterwards, the ML model is applied to these new points to predict the outcomes (scores or probabilities to a certain class). Consequently, a new dataset is created which includes the permutated data points of the features that are close to the original observation and the corresponding predictions (scores).\par
Afterwards, the network model is estimated to explain and interpret the influential features of the predicted outcome, the important interconnections among the features and the significant path route of the model. The proposed algorithm introduces the undirected graphical models as the most well-known frameworks for constructing a network model revealing the straightforward relationships between observable features. Particularly, when data follow a multivariate normal distribution, the produced model is called Gaussian graphical model (GGM; \cite{lauritzen1996graphical}) and belongs to a generalized class of statistical models named pairwise Markov random fields (PMRF; \cite{koller2009probabilistic}). The produced GGM estimates can be standardized, visualized and easier interpreted as partial correlation coefficients \cite{borsboom2013network,MCNALLY201695}. In particular, partial correlation coefficients fluctuate between -1 and 1 and reveal the remaining linear dependency among two variables, after conditioning on all other variables in the dataset \cite{epskamp2018tutorial}. \par

Partial correlations can be directly computed from the inverse of a variance–covariance matrix in which each element represents a weight-edge indicating the strength of connection between two nodes \cite{epskamp2018tutorial}. Let’s assume that $Y^T=[Y_1Y_2…Y_p]$ represents the response vector of a random subject of p features. Supposing that y vector is centred and follows a multivariate normal density with some p×p variance-covariance matrix $\Sigma$, $Y\sim N_p$  (0,$\Sigma$). The partial correlation coefficients are estimated by calculating the inverse of $\Sigma$ (known also as a precision matrix), K=$\Sigma^{-1}$. The element $k_{ij}$ can be standardized to derive the partial correlation coefficient among variables $Y_i$ and $Y_j$ after conditioning on all other variables in $Y$, $Y_{-(i,j)}$, \cite{lauritzen1996graphical}, $Cor(Y_i,Y_j|Y_{-(i,j)})=\frac{-k_{ij}}{(\sqrt{(k_{ii} )} \sqrt{(k_{jj} )}}$.

If the partial correlation coefficient is exactly zero, this signifies a conditional independence between two variables after controlling for all other variables in the model and therefore, no edge is drawn between these two nodes.\par
However, in practice, if two features are conditionally independent, small partial correlations (close to zero) are estimated which are called spurious or false positives \cite{costantini2015state}. In order to control the spurious connections in the precision matrix, a statistical regularization technique originating in the field of machine learning is applied. This technique is known as ‘least absolute shrinkage and selection operator’ (lasso; \cite{tibshirani1996regression}) which shrinks small partial correlation coefficients exactly to zero. The graphical lasso is a regularization framework for estimating the covariance matrix $\Sigma$ under the assumption that precision matrix $K$ is sparse \cite{meinshausen2006high}. The graphical lasso problem maximizes the penalized ($l_1$-regularized) log-likelihood:

\begin{equation} \label{eq1}
maximize_{K > 0} f(K) :=  \log{det(K)} - (SK) - \lambda \lVert \mathbf{K} \rVert_1
\end{equation}

where S denotes the sample covariance matrix, $\lambda$ (lambda) is a nonnegative tuning parameter controlling the amount of $l_1$ shrinkage and $\lVert \mathbf{K} \rVert_1$ is the $l_1$ norm (the sum of the absolute values of the elements of $\Sigma{^{-1}}$). Since $\lambda$ regulates the sparsity of the network, various values of $\lambda$ provide different network structures \cite{zhao2006model} indicating that a group of networks ranging from a fully connected network ($\lambda_{min}$) to an empty network ($\lambda_{max}$) are estimated. Typically, a logarithmically spaced range of tuning parameters in which $\lambda_{min}$=R$\lambda_{min}$ where R = 0.01 by default. Subsequently, the network that minimizes the Extended Bayesian Information Criterion (EBIC; \cite{drton2004model}) is characterized as optimal network signifying that fits better into the data. As shown in (2), the EBIC adds an extra penalty, the hyperparameter Y (gamma) to control the model complexity \cite{foygel2010extended} and it is set manually from 0 to 0.5, L indicates the log-likelihood, n the sample size, E the number of non-zero edges and p the number of nodes.

\begin{equation} \label{eq2}
EBIC =-2L + E \log{n} + 4\gamma E \log{p}  
\end{equation}

It is conducted that the combination of lasso regularization with EBIC model selection provides the true network structure, particularly when a sparser network is estimated (Epskamp and Fried, 2018; Foygel and Drton, 2010). A p×p undirected weighted matrix is returned in which the conditional dependent relationships among variables are presented and stronger connections among features are identified. To explain the features’ importance on the predicted outcome, we rank the features that are directly connected to the predicted feature. Finally, the path that traverses the most important features is computed from the end node (the predicted feature). Specifically, it searches which node is highly connected to the end node, then the selected node searches at its neighborhood the node with the strongest connection. The algorithm stops when i) all nodes of the model are traversed or ii) the edge- weight among node i and node j ($w_{ij}$) is lower than a specific value (by default, it stops when $w_{ij}<0.1$). For the sake of simplicity, the steps of gLIME algorithm are summarized in Pseudocode I.

\begin{table}[!htb]
\caption{The gLIME pseudocode}
\centerline{\includegraphics[scale=0.87]{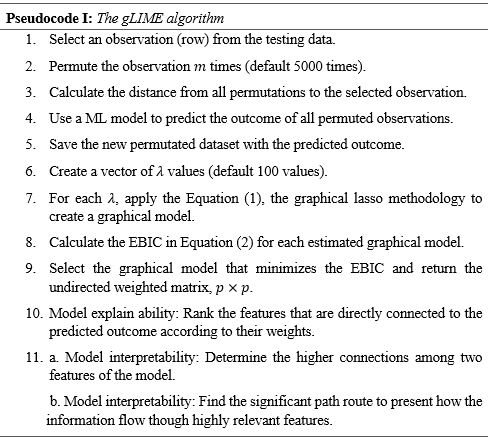}}
\label{table}
\end{table}

\section{Results and Discussion of Results}
\label{headings}

In this study, the gLIME algorithm was applied in two datasets (BIOPSY and OAI) to explain which of the features influenced significantly the predicted output, understand better the estimated graph model by interpretating the strongest connections among the features and indicate the significant path in the graph presenting how the predicted output could be indirectly affected by other features of the network. Every dataset was divided into the training and testing dataset. From the testing data, four random observations were selected and for each observation, ten permutated datasets were generated including 5000 permutated observations. In total, 120 permutated datasets were produced, forty for each problem. These permutations were necessary as well to confirm the stability of the gLIME algorithm compared to the results of LIME. \par
Specifically, the BIOPSY dataset includes biopsies of breast tumors of 699 patients (\url{https://github.com/cran/MASS/blob/master/data/biopsy.rda}). Each of the nine attributes ($V_1$: Clump thickness, $V_2$: Uniformity of cell size, V3: Uniformity of cell shape, $V_4$: Marginal adhesion, $V_5$: Single epithelial cell size, $V_6$: Bare nuclei, $V_7$: Bland chromatin, $V_8$: Normal nucleoli and $V_9$: Mitoses) has been scored on a scale of 1 to 10, and the outcome was classified in two classes ‘benign’ or ‘malignant’.The OAI dataset (https://nda.nih.gov/oai/) focused on Osteoarthritis problem; specifically, the goals of the OAI are to provide resources to enable a better understanding of prevention and treatment of knee osteoarthritis. For this study, the OAI dataset consists of 3873 patients and forty features were examined; while the outcome was categorized in two categories ‘healthy’ and ‘not healthy’. \par
Table 2 illustrates the graphical lasso estimates of the gLIME algorithm applied on the first permutated dataset of the first observation. As it is observed, the derived matrix is a $10\times 10$ undirected weighted matrix in which the number ten represents the overall number features (nine inputs and one output). Moreover, it was also inferred the strongest conditional associations between two features after conditioning on all other features in the model. The indicative strongest positive conditional associations were identified between the Uniformity of cell size and Uniformity of cell shape ($w_{2,3}$=.594), the Uniformity of cell size and the Single epithelial cell size ($w_{2,5}$=.231) and the Single epithelial cell size and Mitoses ($w_{5,9}$=.178). Thus, the strongest negative conditional associations were detected among input features (Bare nuclei, Clump thickness and Bland chromatin) and the predicted outcome which was classified as ‘benign’ after conditioning on all other features in the model, $w_{6,01}$=-.395, $w_{1,01}$=-.304 and $w_{7,01}$=-.216. The negative conditioning connection signifies the reverse relation between two features after conditioning on all other features in the model. For instance, the $w_{6,O1}$=-.395 determines the smaller the Bare nuclei attribute is, the healthier the patient (higher possibility to be benign cancer) after conditioning on all other observed attributes and vice-versa. Moreover, six out of forty-five connections were distinguished as zero implying the conditional independence among features $w_{1,7}$=$w_{1,8}$=$w_{2,6}$=$w_{3,9}$=$w_{5,01}$=$w_{6,8}$=0.

\begin{table}[h]
\centering
\caption{The weight adjacency matrix using gLIME in the first permutated dataset of the first observation applied on ten features (nine inputs: $V_1$-$V_9$ and one output: $O_1$). }
\centerline{\includegraphics[scale=0.5]{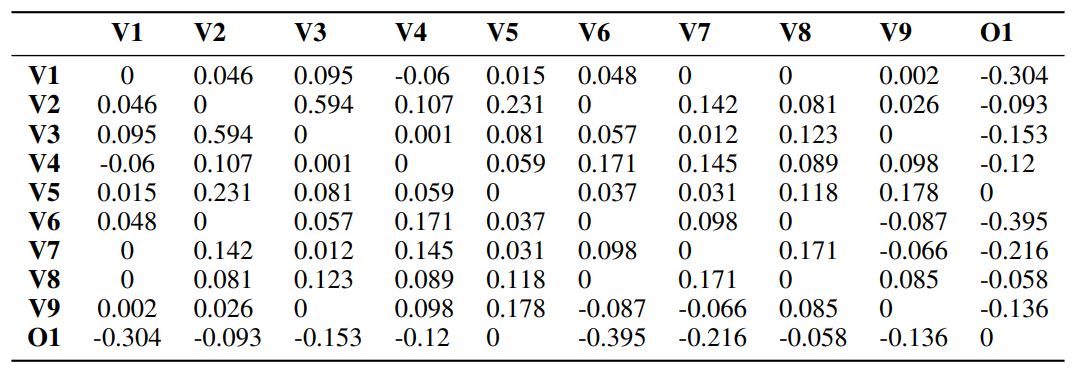}}
\label{table}
\end{table}

\begin{figure}[htb]
    \begin{minipage}[t]{.45\textwidth}
        \centering
        \includegraphics[width=\textwidth]{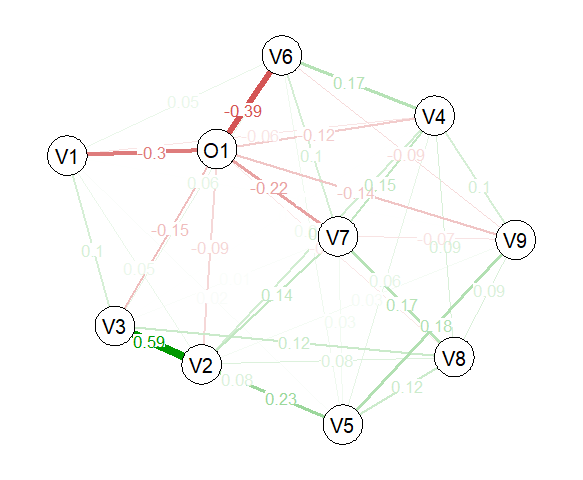}
        \subcaption{Graph structure}\label{fig:1}
    \end{minipage}
    \hfill
    \begin{minipage}[t]{.45\textwidth}
        \centering
        \includegraphics[width=\textwidth]{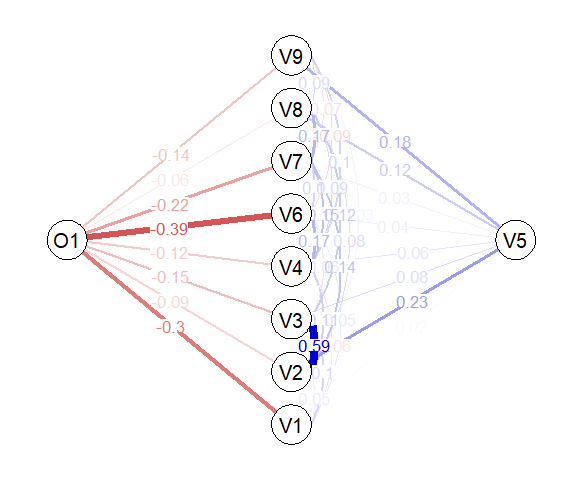}
        \subcaption{Flow diagram.}\label{fig:2}
    \end{minipage}  
    \label{fig:1-2}
    \caption{Visual representation of the regularized partial correlation matrix of Table 1 as a partial network structure. Circles represent the observed features and links represent the regularized partial correlation coefficients among two features after conditioning on all other features of the model. Green/blue and red edges denote positive and negative associations, respectively. Wider and more saturated edges indicate higher strengths among nodes.}
\end{figure}

Since every graph can be described by an adjacency matrix, Figure 1 illustrates the corresponded graphical model of Table 2. Every weight of the graph is colored according to its strength. Specifically, green or blue colors represent positive regularized partial associations and red characterize negative regularized partial associations. Thus, saturated edges define the magnitude of the partial strength among two nodes, i.e. lighter color display weaker regularized partial correlations and darker pigment illustrate higher regularized partial associations. There are many ways to visualize a graph structure. In Figure 1, the weighted matrix of Table 2 has been presented as a) a simple graph in which strongly connected nodes were placed closer reflecting the full picture of the model; while, weakly related nodes were set away from each other and b) a flow diagram in which the node of interest (O1) was placed to the left, while the nodes that were directed connected to the node of interest were placed vertically in a new layer (similar to a neural network structure) and so on, presenting clearly the significant features related to the node of interest.\par

\begin{table}[!htb]
\caption{The ranking positions of the features that are directly linked with the prediction outcome of ten permutated data (1A – 1J) using gLIME and LIME concerning the BIOPSY dataset.}
\centerline{\includegraphics[scale=0.67]{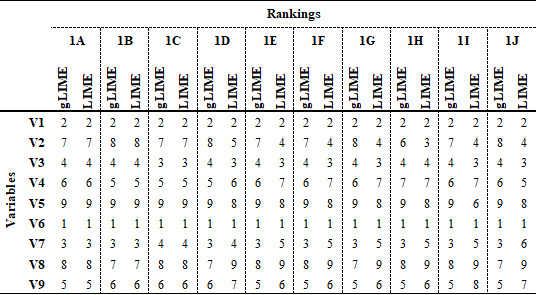}}
\label{table}
\end{table}

\begin{table}[!htb]
\caption{Biopsies of breast tumors problem. The Kendall’s tau coefficients of ten permutated data of the first observation. The $\tau_b$ of gLIME and LIME are highlighted with green and blue color, respectively.}
\centerline{\includegraphics[scale=0.67]{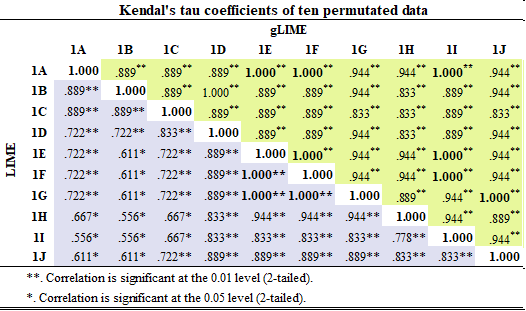}}
\label{table}
\end{table}

The features that were directly connected with the prediction outcome were ranked according to their absolute edge weight. Table 3 includes the ranking results of the first observation using gLIME and LIME over ten permutated data (1A – 1J). Overall, these results indicated that the most important features are V6 and V1; however, after the second position, small differences were observed. To compare the stability and the consistency of gLIME and LIME, Kendall’s tau coefficient ($\tau_b$) was applied \cite{kendall1948rank}. Table 4 illustrates Kendall’s tau coefficients to determine the relationships among the permutated data of the first observation towards gLIME and the realationships ($\tau_b$) of LIME, respectively. The average ($\tau_b$ )  correlation of gLIME highlighted with green color, indicated that there was a very strong, positive relationship that was statistically significant (gLIME: $\tau_b$=.934). On the other hand, the average Kendall’s correlation of LIME depicted with a blue color, determined that there was a strong, positive relationship that was statistically significant (LIME: $\tau_b$=.819). Consequently, the average stability of the first observation using gLIME was higher 93.4\% comparing to LIME (81.9\%). \par
For the sake of completeness, Table 5 summarizes the robustness of the selected observations (1st, 6th, 100th and 600th). Specifically, forty permutated datasets were generated, ten permutated datasets per observation. In total, both algorithms presented high robustness in all observations (over 81.9\%); but, the average stability of each observation in gLIME revealed that the results were less sensitive comparing to LIME. Thus, in the last column of Table 5, the average stability of gLIME has shown higher robustness coefficients over the selected observations in BIOPSY dataset in which a small number of features were tested.In addition, the corresponded summarized stability coefficients of the OAI problem applied on four random observations (1st, 10th, 100th and 200th) using forty permutated datasets (Table 6). The results revealed that gLIME performed better in terms of robustness since the average Kendall’s correlation coefficients were exceeded in all of the selected observations (69.4\% - 81.3\%) comparing to LIME stability results (45.5\% - 58.1\%). Moreover, these results highlighted the inefficiency of LIME regarding the consistency estimates when the number of features were increased. 

\begin{table}[h]
\centering
\caption{Summarized stability coefficients of four observations which were selected from the BIOPSY testing set using gLIME and LIME algorithm. }
\centerline{\includegraphics[scale=0.5]{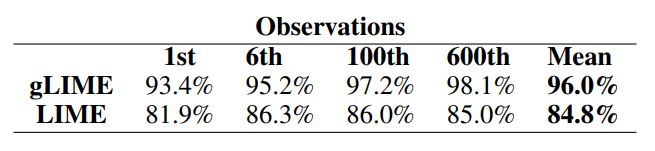}}
\label{table}
\end{table}

\begin{table}[h]
\centering
\caption{Summarized stability coefficients of four observations which were selected from the OAI testing set using gLIME and LIME algorithm studying forty features. }
\centerline{\includegraphics[scale=0.5]{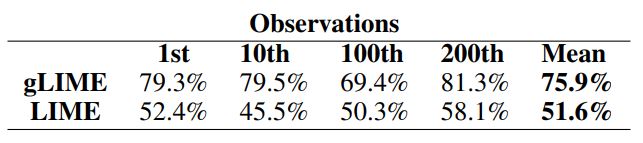}}
\label{table}
\end{table}

Another advantage of gLIME was the ability to interpret how the undirected features were able to influence the prediction of outcome. Figure 2 visualizes the produced graphical gLIME explanations for the OAI problem including forty features. Specifically, the network of the first permutated dataset (100A) of the 100th observation was depicted. For visualization purposes, very small weight edges (w$<$0.1) were excluded from the graph in order to identify easier the significant path routes. \par

The following remarks could be drawn from Figure 2: (i) The classification output (O1) is strongly affected by variables V5 and V33. Green lines reveal a positive effect (an increase in those variables pushes O1 to increase as well) (ii) V5 is also positively affected by V36 that is also similarly correlated with V34. Therefore a path of positive correlations is identified between the variables V34, V36, V5 and the output O1. (iii) Even there is no direct correlation between V34 and the output O1 (as shown in Figure 3 in both LIME and gLIME bar graphs) the produced graph helps the user to identify that there is indirect relationship between them. (iv) There are variables that are strongly correlated to each other but have no (or minor) effect to the output. (v) This descriptive graphical presentation of the feature importance values and their interconnections enhance the users’ understanding of the rationale behind the decision making mechanism of the trained black box model.\par

\begin{figure}[h]
\caption{The graphical representation of the first permutated dataset of the 100th observation. For the sake of simplicity, weights under 0.1 were eliminated from the graph.}
\centerline{\includegraphics[scale=0.85]{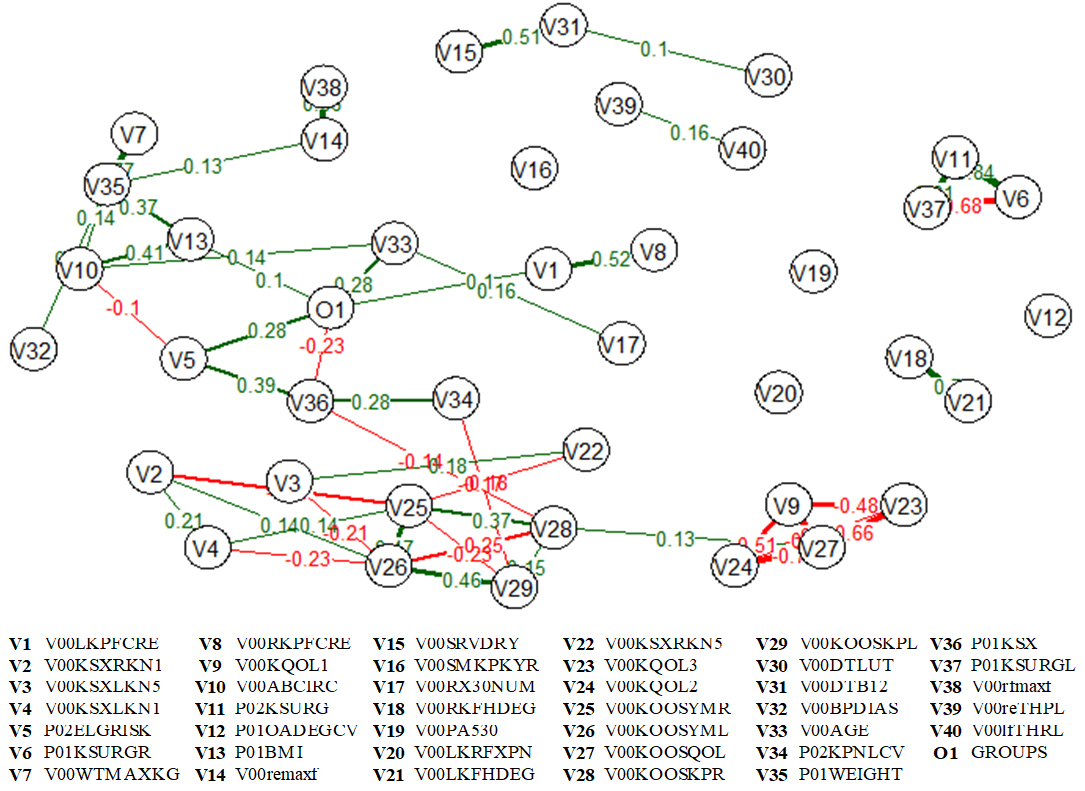}}
\label{fig}
\end{figure}

Figure 3 shows the barplot explanations of both LIME and gLIME. The most important features that affect the predicted outcome are placed at the top of the list. Blue and orange color show positive and negative influence of each feature with the output feature (O1). Similar ranking are produced by both approaches however due to regularized partial correlations of gLIME, the effect of the last seven features is significantly lowered compared to LIME.  

The results of our proposed algorithm (gLIME) are showing superiority in terms of stability  and interpretability compared to LIME in tabular data; however, more research on this topic needs to be undertaken. Therefore, we are planning to apply gLIME to other types of data, such as text and images. Moreover, further research should be done to investigate the robustness of the predicted outcomes when more features (over 100) are included. Finally, introducing causality in the produced graphs is within our future scientific interests that will lead to more meaningful directed graphical explanations.

\begin{figure}[h]
\centering
\caption{The barplot explanations of the model-agnostic algorithms gLIME (left) and LIME (right). }
\centerline{\includegraphics[scale=0.5]{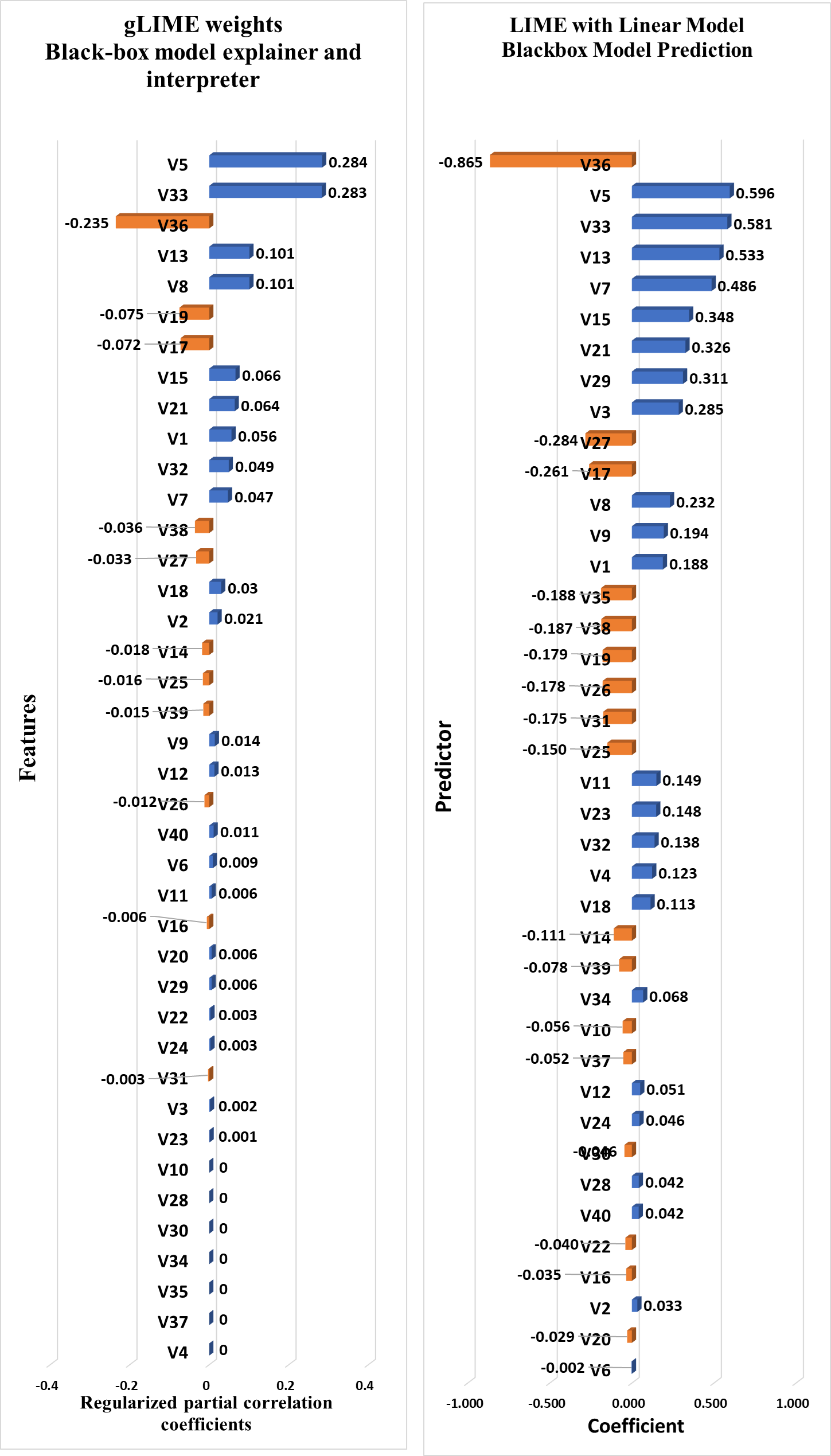}}
\label{fig}
\end{figure}

\section{Conclusions}

gLIME is a novel graphical model-agnostic explainability methodology that goes beyond the current state-of-the-art of XAI methods. It incorporates graph theory to identify the most relevant features that affect the black box model’s output, highlight hidden but important relationships among features quantifying the strength of the relationships and determine significant path routes presenting how the information flows from one node to the end node (the predicted output). This enhanced graphical visualization of the produced explanations can increase significantly the user’s understanding of the inner workings and the reasoning behind the ML models’ decision-making process. Apart from being interpretable, gLIME was also proved to be more robust and consistent compared to LIME on an extensive experimentation that included two well-known classification datasets. gLIME’s informative and graphically given explanations that could unlock black boxes contributing to the development of robust and trustworthy AI-empowered systems.

\section{Acknowledgement}
This research was funded by the European Community’s H2020 Programme, under grant agreement No. 957362 (XMANAI).

\bibliographystyle{unsrtnat}
\bibliography{references}  

\end{document}